\documentclass[sigconf]{acmart}

\AtBeginDocument{%
  \providecommand\BibTeX{{%
    \normalfont B\kern-0.5em{\scshape i\kern-0.25em b}\kern-0.8em\TeX}}}

\copyrightyear{2020}
\acmYear{2020}
\setcopyright{acmcopyright}\acmConference[MM '20]{Proceedings of the 28th ACM International Conference on Multimedia}{October 12--16, 2020}{Seattle, WA, USA}
\acmBooktitle{Proceedings of the 28th ACM International Conference on Multimedia (MM '20), October 12--16, 2020, Seattle, WA, USA}
\acmPrice{15.00}
\acmDOI{10.1145/3394171.3413857}
\acmISBN{978-1-4503-7988-5/20/10}

\usepackage[ruled, vlined, linesnumbered]{algorithm2e}
\usepackage{subcaption}
\usepackage{tabularx}

\begin{document}
\fancyhead{}

\title{Self-Play Reinforcement Learning for Fast Image Retargeting}

\author{Nobukatsu Kajiura}
\email{n\_kajiura@hal.t.u-tokyo.ac.jp}
\affiliation{%
  \institution{The University of Tokyo}
  \state{Tokyo}
  \country{Japan}}

\author{Satoshi Kosugi}
\email{kosugi@hal.t.u-tokyo.ac.jp}
\affiliation{%
  \institution{The University of Tokyo}
  \state{Tokyo}
  \country{Japan}}

\author{Xueting Wang}
\email{xt\_wang@hal.t.u-tokyo.ac.jp}
\affiliation{%
  \institution{The University of Tokyo}
  \state{Tokyo}
  \country{Japan}}

\author{Toshihiko Yamasaki}
\email{yamasaki@hal.t.u-tokyo.ac.jp}
\affiliation{%
  \institution{The University of Tokyo}
  \state{Tokyo}
  \country{Japan}}


\begin{abstract}
    In this study, we address image retargeting, which is a task that adjusts input images to arbitrary sizes. In one of the best-performing methods called MULTIOP, multiple retargeting operators were combined and retargeted images at each stage were generated to find the optimal sequence of operators that minimized the distance between original and retargeted images. The limitation of this method is in its tremendous processing time, which severely prohibits its practical use. Therefore, the purpose of this study is to find the optimal combination of operators within a reasonable processing time; we propose a method of predicting the optimal operator for each step using a reinforcement learning agent. The technical contributions of this study are as follows. Firstly, we propose a reward based on self-play, which will be insensitive to the large variance in the content-dependent distance measured in MULTIOP. Secondly, we propose to dynamically change the loss weight for each action to prevent the algorithm from falling into a local optimum and from choosing only the most frequently used operator in its training. Our experiments showed that we achieved multi-operator image retargeting with less processing time by three orders of magnitude and the same quality as the original multi-operator-based method, which was the best-performing algorithm in retargeting tasks.
\end{abstract}





\maketitle

\section{Introduction}
Image retargeting, which is a task of adjusting input images into arbitrary sizes, has been actively studied owing to the diversity in display devices and the versatility in the media sources of images.
In image retargeting, it is important to generate natural results, while retaining important objects/regions.
Nevertheless, it is difficult to achieve this with simple operations, such as cropping or uniform scaling.
Although some content-aware retargeting methods have been proposed~\cite{avidan2007seam, rubinstein2008improved, han2010optimal, frankovich2011enhanced, liu2005automatic, gal2006feature, wolf2007non, wang2008optimized, krahenbuhl2009system}, using a single retargeting operator will not succeed in all cases or for all sizes.
In this study, we apply a multi-operator image retargeting by utilizing several retargeting operators and appropriately combining them to obtain better results that will be tuned for each image.

\begin{figure}[t]
  \centering
  {\tabcolsep=0mm
  \begin{tabular}{l l r}
  &&\\
      \begin{minipage}{0.49\hsize}
          \begin{center}
          \includegraphics[width=\hsize]{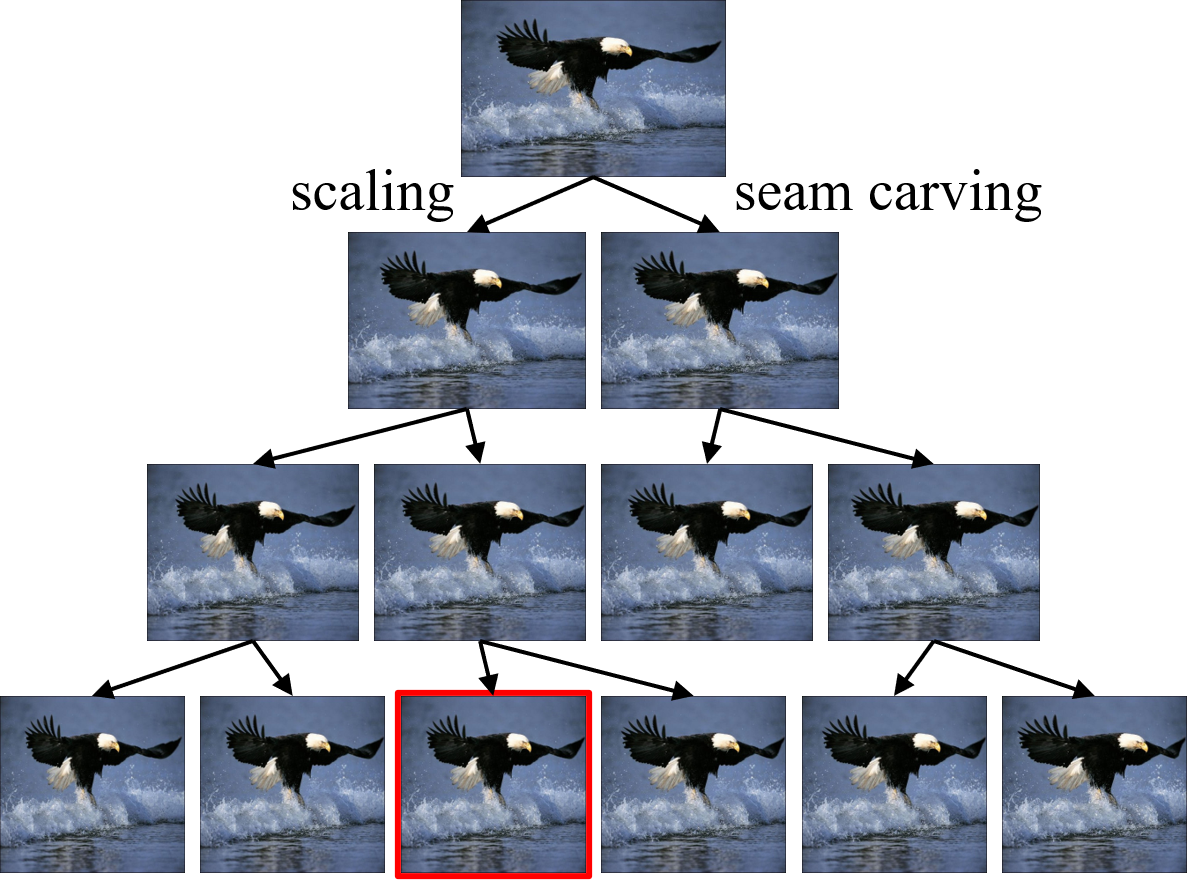}
          \subcaption{MULTIOP~\cite{rubinstein2009multi}}
          \end{center}
      \end{minipage}\hspace{0.1ex}
      &&\hspace{0.1ex}
      \begin{minipage}{0.49\hsize}
          \begin{center}
          \includegraphics[width=\hsize]{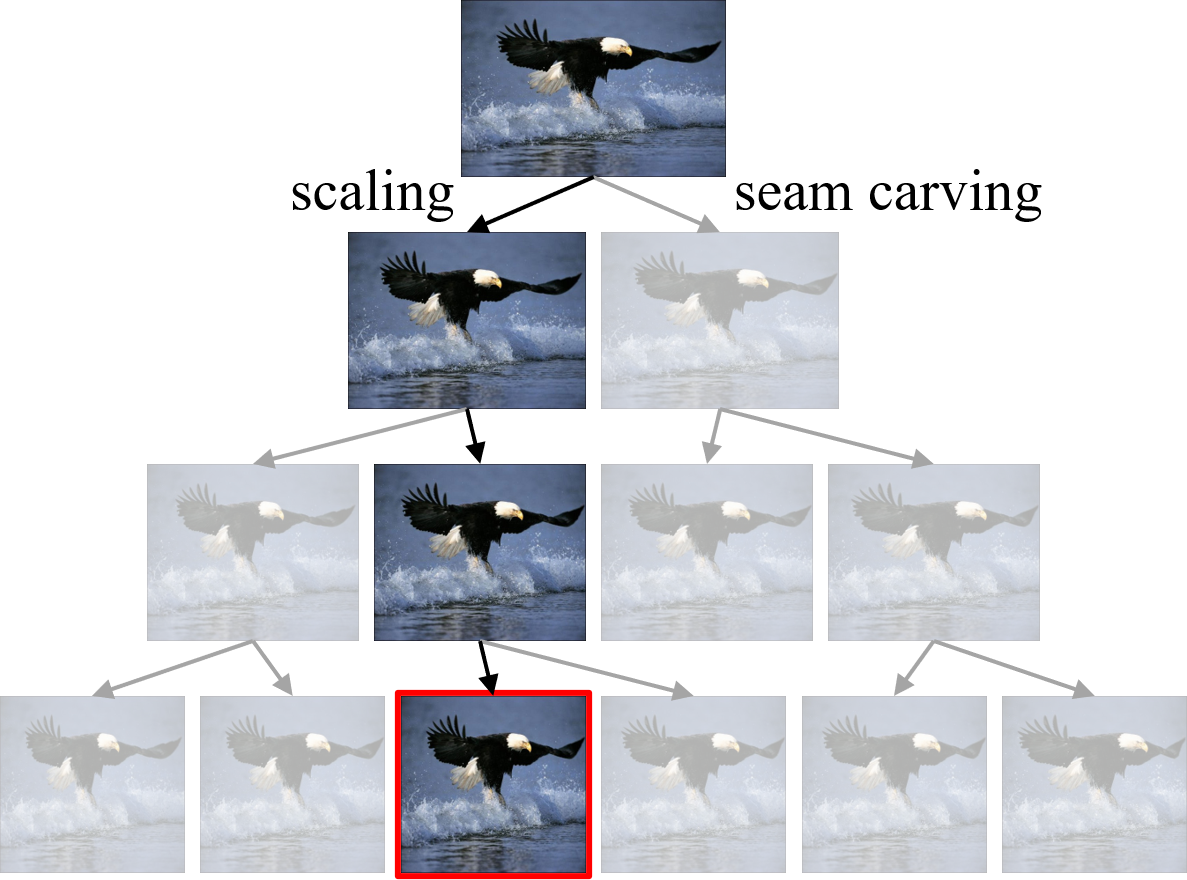}
          \subcaption{Ours}
          \end{center}
      \end{minipage}\\
  \end{tabular}
  }
  \vspace{-2ex}
  \caption{Comparison of how to search for the optimal retargeted image by multi-operator image retargeting methods. For simplicity, the figure shows a three-step case with two kinds of operators (scaling and seam carving). MULTIOP~\cite{rubinstein2009multi} combines multiple retargeting operators and generates retargeted images at each stage. Whereas, in our method, the optimal retargeting operator is predicted for each step by a reinforcement learning agent.}
  \label{fig:compare_multiop}
\end{figure}

In a previous method using multi-operators, Rubinstein et al.~\cite{rubinstein2009multi} proposed an image-to-image distance measure called Bi-Directional Warping (BDW), and the optimal combination of retargeting operators was searched for using dynamic programming.
Although they achieved a better performance than other retargeting approaches, their approach had a critical limitation: a huge computational time.
This is because retargeted images using multiple operators had to be generated at each stage (Figure~\ref{fig:compare_multiop}(a)).
In contrast, we propose a high-speed multi-operator image retargeting by predicting the optimal retargeting operator step by step.
We achieve this by using a reinforcement learning agent instead of generating multiple images to search for retargeting operator combinations (Figure~\ref{fig:compare_multiop}(b)).
By improving the efficiency of searching for the appropriate retargeting operators via reinforcement learning, fewer images are generated and the computational time is drastically reduced.

The purpose of the agent is to find retargeting operators that can minimize the distance between the original image and the retargeted image as much as possible.
When applying reinforcement learning to this search, we find two problems and propose the following solutions for each.
Firstly, as the dynamic range of distance (the BDW score) varies greatly depending on the image content, the distance cannot be directly used as a reward for training. Besides, we also find that predicting the BDW score via neural networks is extremely difficult.
To solve this problem, we propose a self-play-based reward.
By making an agent play against its copy and calculating the reward based on the victory or defeat, the agent can be trained based not on the absolute BDW score but the relative score between them.
Secondly, simply using victory or defeat as a reward to the agent often leads to overfitting in which only one or two actions are selected.
The retargeted images are then often worse than the results of MULTIOP~\cite{rubinstein2009multi}.
This is because the chance of victory increases by just picking a relatively strong action.
In order to solve this problem, we propose to dynamically change the loss weight of each action.
By changing the weight of the loss according to the frequency of the action selected, the selection probabilities of the relatively strong action and the relatively less frequently used action are evaluated equally.
As a result, the agent can correctly learn the optimal action.

Our experiments show that our method is faster by three orders of magnitude than MULTIOP~\cite{rubinstein2009multi}.
Furthermore, we also show that our method can achieve the same image quality as MULTIOP~\cite{rubinstein2009multi} and that our image quality is better than those of the other state-of-the-art approaches.

Our main contributions are summarized as follows:
\begin{itemize}
  \item We propose a reinforcement-learning-based method that can achieve ultra-fast multi-operator image retargeting.

  \item We show that a self-play-based reward can be insensitive to the large variance in the distance measure.
  
  \item We propose to dynamically change the loss weight of each action so that multiple operators could be evaluated and selected in balance to avoid overfitting.

  \item Experiments show that our method achieves multi-operator image retargeting faster by three orders of magnitude, and achieves the same image quality as MULTIOP~\cite{rubinstein2009multi} according to the user study.
\end{itemize}

\section{Related Work}
% In this section, we first introduce image retargeting approaches and then review reinforcement learning methods for image processing.

\subsection{Image Retargeting}
For image retargeting, single-operator-based methods including hand-crafted techniques, deep-learning-based methods, and multi-operator-based methods are introduced.
The key factor in image retargeting is how to suppress the loss and distortion of content.
% Therefore, content-aware image retargeting approaches often employ importance maps that are created from original images.

A typical method for image retargeting is seam carving~\cite{avidan2007seam}. This method finds the optimal seam according to the image energy map via dynamic programming, and continuously removes seams to change the image size.
Rubinstein et al.~\cite{rubinstein2008improved} introduced a new energy map and a graph-cut approach and achieved temporal-consistency-aware video retargeting.
Han et al.~\cite{han2010optimal} found multiple seams simultaneously with region smoothness and seam shape prior to using a 3-D graph-theoretic approach.
Frankovich and Wong~\cite{frankovich2011enhanced} introduced the absolute-energy cost function, which penalized seam candidates that crossed areas of local extrema.

Retargeting approaches based on warping were also reported.
Liu and Gleicher~\cite{liu2005automatic} used a non-linear fisheye-view warp that emphasized important regions while shrinking others.
Gal et al.~\cite{gal2006feature} designed a warping technique to preserve user-specified features by constraining their deformation to be a similarity transformation.
Wolf et al.~\cite{wolf2007non} introduced a non-homogenous mapping of video frames and retargeted videos by warping.
Wang et al.~\cite{wang2008optimized} proposed scale-and-stretch warping, which distributed the distortion in all spatial directions, thus utilizing the available homogeneous regions to suppress the overall distortion.
Kr{\"a}henb{\"u}hl et al.~\cite{krahenbuhl2009system} proposed streaming video, which is a non-uniform and pixel-accurate warp to the target resolution.
% , which considered automatic features as well as interactively-defined features.

Recently, deep-learning-based methods have been proposed.
Cho et al.~\cite{cho2017weakly} proposed a weakly- and self-supervised learning model~(WSSDCNN)~that learns the shift map of each pixel in the input and output images.
Tan et al.~\cite{tan2019cycle} proposed an unsupervised learning model, Cycle-IR, which learned the forward and reverse mapping of input and output images.
Lee et al.~\cite{lee2020object} used object detection and object tracking with a deep neural network to enable consistent video retargeting.
The deep-learning-based methods have a great advantage because their inference time can be very short.
However, they are still inferior to hand-crafted algorithms because of the distortions in the resultant images.

Rubinstein et al.~\cite{rubinstein2009multi} claimed that using multiple retargeting operators for resizing images is often better than using a single one and proposed a method called MULTIOP, which searched for a suitable retargeting method for each image by combining multiple retargeting methods step by step.
They included cropping, scaling, and seam carving into their retargeting operators, generated multiple retargeted images using dynamic programming, and searched for the optimal combination of retargeting operators for the original image.
To evaluate the distance between original and retargeted images, they defined a new image similarity measure, called BDW.
In a user study using the RetargetMe dataset~\cite{rubinstein2010comparative}, MULTIOP achieved the highest rating together with streaming video~\cite{krahenbuhl2009system}.
Because the combination of multiple retargeting operators achieved high performance, various improvements have been proposed.
Zhang et al.~\cite{zhang2017multi} kept the quality of visually important objects by stretching original images in both vertical and horizontal directions and then applying seam carving and scaling.
Song et al.~\cite{song2018photo} proposed deep-learning-based multi-operator image retargeting by learning the proportion of cropping, scaling, and seam carving operators.
Compared to these two methods, we do not fix the order of retargeting operators, so we can explore a wider space of operator combinations.
Zhou et al.~\cite{zhou2020weakly} first applied reinforcement learning to find the optimal combination of multiple operators and used a semantic and aesthetic reward.
However, we find that directly applying reinforcement learning could not use the BDW score as a reward.
To solve this problem, we propose a self-play reinforcement learning architecture.
The proposal can still exploit the advantage of BDW while aesthetic reward can be employed as well.

\begin{figure*}[t]
  \includegraphics[width=\textwidth]{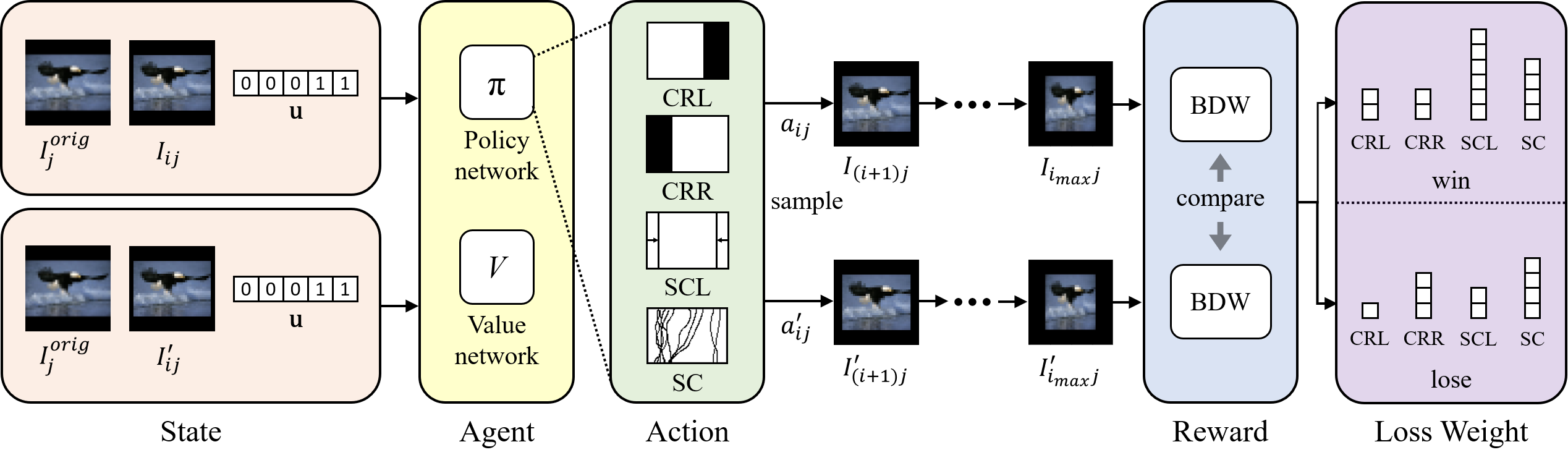}
  \caption{Illustration of the self-play reinforcement-learning-model architecture. The agent repeats receiving of the observation composed of the original image, the current retargeted image, a one-hot vector representing the number of steps to the end of the episode, and sampling two actions of the self and the opponent according to the policy output. Each current retargeted image is updated by the retargeting operator corresponding to the sampled action. At the end of the episode, the agent receives a reward based on the victory or defeat of the BDW score. The loss weight of the value network and the policy network are changed according to the number of times each action is selected in case of victory or defeat.}
  \label{fig:architecture}
\end{figure*}

\subsection{Reinforcement Learning for Image Processing}
In recent years, reinforcement learning has been applied to image processing applications.
Cao et al.~\cite{cao2017attention} proposed a super-resolution method for facial images by letting the agent choose the local region to be enhanced.
% Cao et al.~\cite{cao2017attention} proposed a super-resolution method for facial images by letting the agent choose the local region that should be enhanced.
Park et al.~\cite{park2018distort} used Deep Q-Network~\cite{mnih2015human} for color enhancement by iteratively choosing the image manipulation action.
Hu et al.~\cite{hu2018exposure} proposed a photo retouching method for RAW images by choosing the image manipulation filter.
Yu et al.~\cite{yu2018crafting} proposed an image restoration method by selecting a toolchain from a toolbox.
Furuta et al.~\cite{furuta2019fully, furuta2019pixelrl} proposed a fully convolutional network that allowed agents to perform pixel-wise manipulations for image denoising, image restoration, and color enhancement.
Ganin et al.~\cite{ganin2018synthesizing} used an adversarially trained agent for synthesizing simple images of letters or digits using a non-differentiable renderer.
Kosugi and Yamasaki~\cite{kosugi2019unpaired} used reinforced adversarial learning for photo enhancement utilizing unpaired training data.
Li et al.~\cite{li2018a2} proposed Aesthetics Aware Reinforcement Learning~(A2-RL), which improved the aesthetic quality of images via image cropping.
The agent iteratively chose the region of the cropping window to maximize the aesthetic score of the cropped image.

\section{Method}
The purpose of this study is to perform content-aware image retargeting.
In MULTIOP~\cite{rubinstein2009multi}, multiple retargeted images were generated at each step, and the optimal sequence of operations was decided using dynamic programming.
However, the computational time was a big issue.
We propose a method for predicting the optimal retargeting operator step by step using a reinforcement learning agent.
While the conventional method generates multiple images with pruning, the proposed method generates retargeted images using the shortest path~(Figure~\ref{fig:compare_multiop}).

We show the overview of our method in Figure~\ref{fig:architecture}.
We formulate image retargeting as a sequential decision-making process.
The agent interacts with the environment and chooses an action to optimize the target. 
When we denote the original image as $I^{orig}_j$ and the current step number as $i$, the agent first receives the current state $s_{ij}$, which contains $I^{orig}_j$ and the current retargeted image $I_{ij}$.
Note that the retargeted image in the first step is the same as the original image ({\it i.e., $I_{0j} = I^{orig}_j$}).
Then, the agent samples the action $a_{ij}$ from the action space according to the probability distribution of the learned policy.
% calculated from the state.
Based on the selected action $a_{ij}$, the current retargeted image $I_{ij}$ is updated using a retargeting function $F$, that is, $I_{(i+1)j} = F(I_{ij}, a_{ij})$.
This new image is used to make the new state $s_{(i+1)j}$, and the agent repeats the action sampling based on $s_{(i+1)j}$.
This sequential decision-making process is repeated $i_{max}$ times where $I_{i_{max}j}$ is used as the final retargeting result.
As the proposed method does not need to generate multiple images by dynamic programming, the sequential process is faster than MULTIOP~\cite{rubinstein2009multi}.
According to the evaluation of $I_{i_{max}j}$, the agent receives a reward $R_{i_{max}j}$ at the end of the episode, and based on $R_{i_{max}j}$, the reward for the action of the agent at each step is defined as $R_{ij} = R_{i_{max}j} \times \gamma^{i_{max}-i}$, where $\gamma$ is the discount factor.

As a reinforcement learning algorithm, we use the asynchronous advantage actor-critic (A3C)~\cite{mnih2016asynchronous}, which consists of two networks.
The first one is a value network $V(s_{ij}; \theta_v)$, which estimates the value of the current state.
The loss function to optimize the network parameter $\theta_v$ is defined so that $V(s_{ij}; \theta_v)$ can predict the reward,
\begin{equation}
  L^v_{ij} = (R_{ij} - V(s_{ij}; \theta_v)) ^ 2 / 2.
\end{equation}
The second network is a policy network $\pi(a_{ij} | s_{ij}; \theta_\pi)$, which outputs the probability of each action.
The network parameter $\theta_\pi$ is optimized to minimize the following loss function,
\begin{equation}
  L^\pi_{ij} =  - \log\pi(a_{ij} | s_{ij}; \theta_\pi) (R_{ij}-V(s_{ij}; \theta_v)) - \beta H(\pi(s_{ij}; \theta_\pi)),
\end{equation}
where $H$ is a function that calculates entropy, which encourages the agent to explore and prevents convergence to a local optimum.
By minimizing $L^\pi_{ij}$, the policy network $\pi(a_{ij} | s_{ij}; \theta_\pi)$ is trained to maximize the expected reward.

In the following sections, we describe in detail the state and action spaces, the reward, and the training loss of our framework.
% In the following sections, we give detailed explanations of the state and action spaces, the reward, and the training loss of our framework.

\subsection{State and Action Spaces}
In A3C, the agent determines the action according to the policy output calculated from the current state at each step.
The state contains the observation from the environment.
In a sequential decision-making process, the current state can be represented as $s_{ij} = \{o_{0j},o_{1j},\cdots,o_{(i-1)j},o_{ij}\}$, where $o_{ij}$ is the current observation of the agent.
The historical experience is usually important for future decision-making because a human’s decision-making process considers not only the current observation but also historical experience.
To memorize the historical observations, we use an LSTM unit following the A2-RL model~\cite{li2018a2}.
% To memorize the historical observations $\{o_{0j},o_{1j},\cdots,o_{(i-1)j},o_{ij}\}$, we use an LSTM unit following the A2-RL model~\cite{li2018a2}.
In our model, the current observation $o_{ij}$ consists of the original image $I^{orig}_j$, the current retargeted image $I_{ij}$, and a one-hot vector ${\bf u}$ representing the number of steps to the end of the episode.

As for the action space, we define the agent's actions as selecting a retargeting operator and applying that operator to the image.
% As for the action space, we define the actions of the study as selecting a retargeting operator and applying that operator to the image.
In our model, left cropping (CRL), right cropping (CRR), scaling (SCL), and seam carving (SC)~\cite{rubinstein2008improved} are used as retargeting operators.
We let $a_{ij}$ take the value of $\{0, 1, 2, 3\}$ and associate each value to each operator.
Note that these retargeting operators are chosen to match those in MULTIOP~\cite{rubinstein2009multi}, and it is easy to add or delete a retargeting operator as needed.
All these actions adjust the image width by 2.5\% of the original image size.
The observation and action space are illustrated in Figure~\ref{fig:architecture} for an intuitional representation.

\subsection{Self-Play-based Reward}
\begin{figure}[t]
  \includegraphics[width=\linewidth]{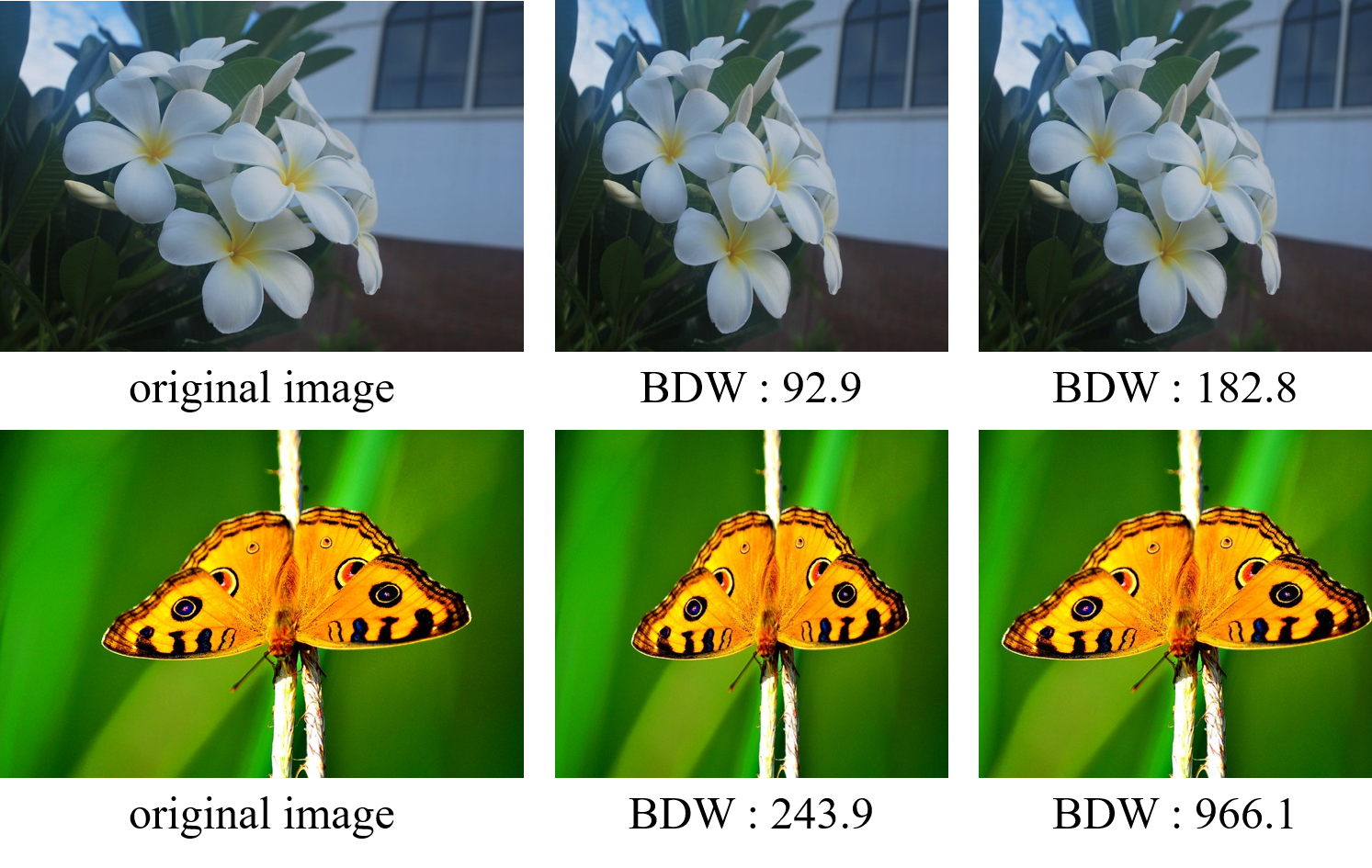}
  \vspace{-5ex}
  \caption{Examples of BDW scores between original and retargeted images. The scale of the BDW score significantly differs for each image. (\textcopyright Renzelle Mae Abasolo, whologwhy)}
  \label{fig:bdw}
\end{figure}

In our reinforcement learning framework, the agent receives a reward according to the evaluation of the retargeted image.
In a previous retargeting method, MULTIOP~\cite{rubinstein2009multi}, they defined a new image similarity measure named BDW and used this measure to evaluate the distance between the original image and the retargeted image.
In this study, we also provided the agent with a reward based on BDW.
Note that other evaluation functions such as aesthetic score~\cite{wang2019aspect} can be used as a reward.
The simplest reward for the agent is the BDW score itself.
However, due to the BDW algorithm, the scale of the BDW score significantly differs for each image~(see Figure~\ref{fig:bdw}) and cannot be approximated by neural networks.
% However, due to the calculation method of the BDW score, the scale of the BDW score significantly differs for each image~(see Figure~\ref{fig:bdw}) and cannot be approximated by neural networks.
If the BDW score is used as a reward as it is, the value cannot be predicted and the reinforcement learning will not proceed normally.

To deal with the large variance in the BDW score, we normalize the evaluation value by self-play reinforcement learning.
Self-play reinforcement learning is a promising method of reinforcement learning.
This method is mainly used for learning game strategies such as chess, shogi, Go, and Othello~\cite {silver2017mastering, silver2018general, van2013reinforcement}.
In this study, we extended a task of image retargeting to ``a game in which players select retargeting operators for the input image, and the victory or defeat is determined by the BDW score.''
In other words, we propose a model in which the agent plays against its copy and receives a reward based on victory or defeat. This self-play-based reward can be insensitive to the large variance in the evaluation value.

The architecture of the self-play reinforcement learning model is illustrated in Figure~\ref{fig:architecture}.
The agent receives two states, $s_{ij}$ and $s'_{ij}$, and samples two actions $a_{ij}$ and $a'_{ij}$ from the action space according to the probability distribution of the policy output $\pi(a_{ij}|s_{ij};\theta_\pi)$ and $\pi(a'_{ij}|s'_{ij};\theta_\pi)$.
Subsequently, the agent executes the sampled actions to update the current retargeted images $I_{ij}$ and $I'_{ij}$ to the new retargeted images $I_{(i+1)j}$ and $I'_{(i+1)j}$, respectively.
The observation of the state and the selection of the action are repeated, and at the end of an episode, the agent receives a reward based on the victory or defeat of the BDW score.
The self-play-based reward $R_{i_{max}j}$ is formulated as:
\begin{equation}
  R_{i_{max}j}=\left\{
  \begin{array}{ll}
    \!\!+1~\left({\rm if~} {\rm BDW}(I^{orig}_j,I_{i_{max}j})\!<\!{\rm BDW}(I^{orig}_j, I'_{i_{max}j})\right)\!\!\!\!\\
    \!\!-1~({\rm otherwise})
  \end{array}.
  \right
  .\label{eq:sp-reward}
\end{equation}
By using the self-play-based reward, the agent considers only the relative BDW score and can deal with the large variance in the BDW score.

\subsection{Frequency-Aware Weighted Loss}
When using a self-play-based reward, we find that the agent often leads to a local optimum in which only a few actions are selected and the retargeted images are often worse than the results of MULTIOP~\cite{rubinstein2009multi}.
This is because the chances of victory are increased by just picking a relatively strong action.
To solve this problem, we propose to dynamically change the loss weight of each action.
Hence, the policy output of the relatively strong action and the relatively weak action are evaluated in balance; we change the loss weight according to the number of times the action is selected.
In every episode, we count how many times each action is selected in the case of winning and losing.
We define the counted results as four-dimensional vectors ${\bf f}^{win}$ and ${\bf f}^{lose}$.
For example, if $a_{ij}$ takes the value of 1 three times, and the agent wins, $f^{win}_1$ increases by three.
We count these numbers over multiple images $I_j$ and we treat these multiple images as a mini-batch ${\mathcal B}$.
When the network parameters are updated, the loss weight $w_{ij}$ is calculated as:
\begin{equation}
  w_{ij}=\left\{
  \begin{array}{ll}
    \!\!min({\bf f}^{win})/f^{win}_{a_{ij}}&({\rm if~}R_{i_{max}j}=+1)\\
    \!\!min({\bf f}^{lose})/f^{lose}_{a_{ij}}&({\rm if~}R_{i_{max}j}=-1)
  \end{array},
\right.
\end{equation}
and gradients with reference to $\theta_\pi$ and $\theta_v$ are accumulated as:
\begin{equation}
d\theta_\pi=\sum_{j=0}^{|{\mathcal B}|-1} \sum_{i=0}^{i_{max}-1} \nabla_{\theta_\pi} w_{ij} L^\pi_{ij}~,
\end{equation}
\begin{equation}
d\theta_v=\sum_{j=0}^{|{\mathcal B}|-1} \sum_{i=0}^{i_{max}-1} \nabla_{\theta_v} w_{ij} L^v_{ij}~,
\end{equation}
where $|{\mathcal B}|$ denotes the batch size of the mini-batch ${\mathcal B}$.
Utilizing this frequency-aware weighted loss, it is possible to avoid falling into a local optimum, in which the relatively strong action is always selected.
We provide the whole training procedure of our reinforcement learning model in Algorithm~\ref{alg:training_procedure}.

\begin{algorithm}[t]
\SetAlgoLined
\SetKwInput{Input}{Input}
\Input{mini-batch of original images $\mathcal{B}$}
$i_{max}$ is sampled from $\{1,...,20\}$\;
${\bf f}^{win}= [0,0,0,0], {\bf f}^{lose}= [0,0,0,0]$\;
\For{$j=0$ \KwTo $|\mathcal{B}|-1$}{
$I_j^{orig}=\mathcal{B}_j$\;
$I_{0j}=I_j^{orig},~I'_{0j}=I_j^{orig}$\;
${\bf u}=[\underbrace{0,...,0}_{20-i_{max}},\underbrace{1,...,1}_{i_{max}}]$\;
\For{$i=0$ \KwTo $i_{max}-1$}{
Make the state $s_{ij}$ from $I^{orig}_j$, $I_{ij}$, and ${\bf u}$\;
Make the state $s'_{ij}$ from $I^{orig}_j$, $I'_{ij}$, and ${\bf u}$\;
Sample $a_{ij}$ and $a'_{ij}$ according to $\pi(a_{ij}|s_{ij};\theta_\pi)$ and $\pi(a'_{ij}|s'_{ij};\theta_\pi)$\;
$I_{(i+1)j} = F(I_{ij}, a_{ij}),~I'_{(i+1)j} = F(I'_{ij}, a'_{ij})$\;
$u_{20-i_{max}+i}=0$\;
}
$R_{i_{max}j}=\left\{
  \begin{array}{ll}
    \!\!+1&({\rm if~} {\rm BDW}(I_j^{orig},I_{i_{max}j})<{\rm BDW}(I_j^{orig},I'_{i_{max}j}))\!\!\!\\
    \!\!-1&({\rm otherwise})
  \end{array}
\right.$\;
\For{$i=i_{max}-1$ \KwTo $0$}{
\eIf{$R_{i_{max}j}==1$}{
$f^{win}_{a_{ij}} += 1$\;
}{
$f^{lose}_{a_{ij}} += 1$\;
}
}
}
$d\theta_\pi=0,~d\theta_v=0$\;
\For{$j=0$ \KwTo $|\mathcal{B}|-1$}{
\For{$i=i_{max}-1$ \KwTo $0$}{
$w_{ij}=\left\{
  \begin{array}{ll}
    \!\!min({\bf f}^{win})/f^{win}_{a_{ij}}&({\rm if~}R_{i_{max}j}==+1)\!\!\!\\
    \!\!min({\bf f}^{lose})/f^{lose}_{a_{ij}}&({\rm if~}R_{i_{max}j}==-1)\!\!\!
  \end{array}
\right.$\;
$R_{ij}= \gamma R_{(i+1)j}$\;
$ d\theta_\pi=d\theta_\pi + \nabla_{\theta_\pi} w_{ij} L^\pi_{ij}$\;
$d\theta_v=d\theta_v + \nabla_{\theta_v} w_{ij} L^v_{ij}~$\;
}
}
Update $\theta_\pi$ with $d\theta_\pi$ and $\theta_v$ with $d\theta_v$\;
\caption{Training procedure of our self-play RL model}
\label{alg:training_procedure}
\end{algorithm}

\begin{figure*}[t]
  \centering
  \begin{tabular}{c c c c c c c c}
      \begin{minipage}{0.21\hsize}
          \begin{center}
          \includegraphics[width=\hsize]{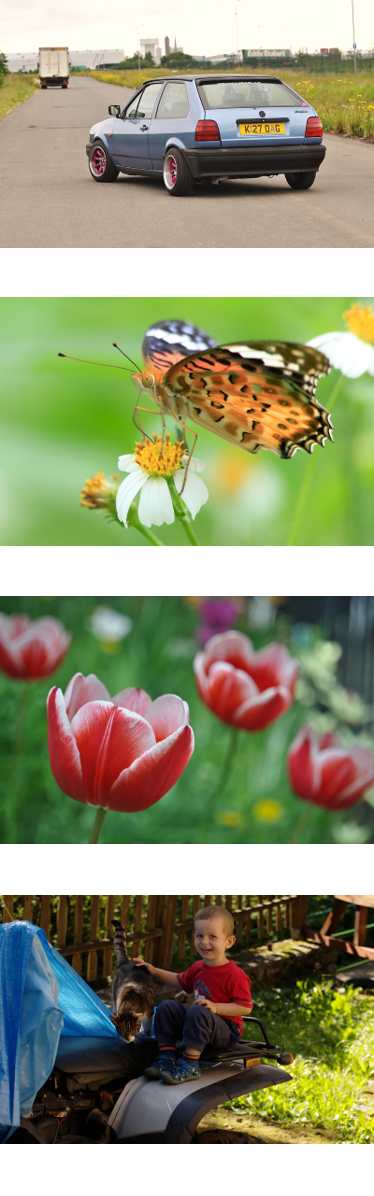}
          \vspace{-5ex}
          \subcaption{\footnotesize original image}
          \end{center}
      \end{minipage}
      \begin{minipage}{0.105\hsize}
          \begin{center}
          \includegraphics[width=\hsize]{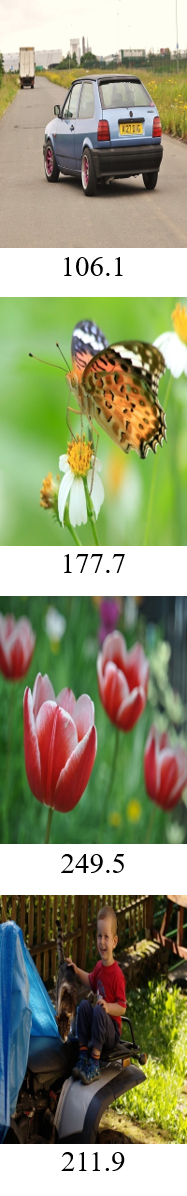}
          \vspace{-5ex}
          \subcaption{\footnotesize SCL}
          \end{center}
      \end{minipage}
      \begin{minipage}{0.105\hsize}
          \begin{center}
          \includegraphics[width=\hsize]{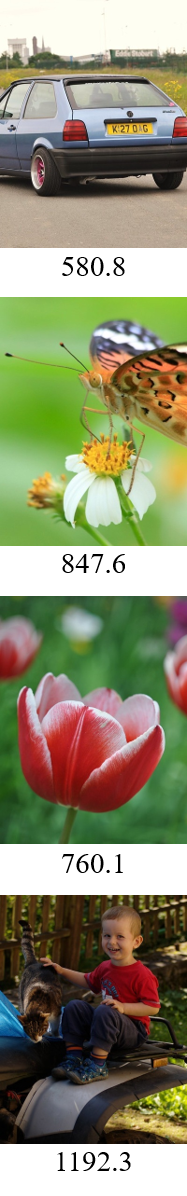}
          \vspace{-5ex}
          \subcaption{\footnotesize GAIC}
          \end{center}
      \end{minipage}
      \begin{minipage}{0.105\hsize}
          \begin{center}
          \includegraphics[width=\hsize]{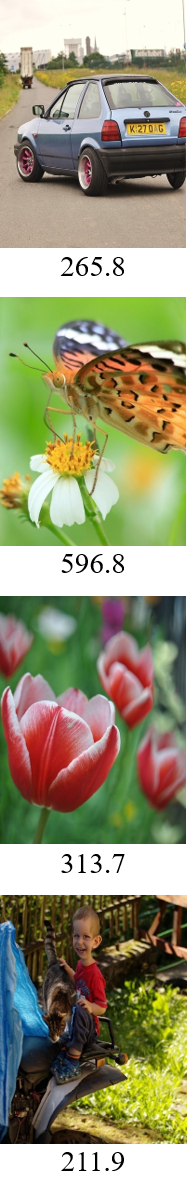}
          \vspace{-5ex}
          \subcaption{\footnotesize SC}
          \end{center}
      \end{minipage}
      \begin{minipage}{0.105\hsize}
          \begin{center}
          \includegraphics[width=\hsize]{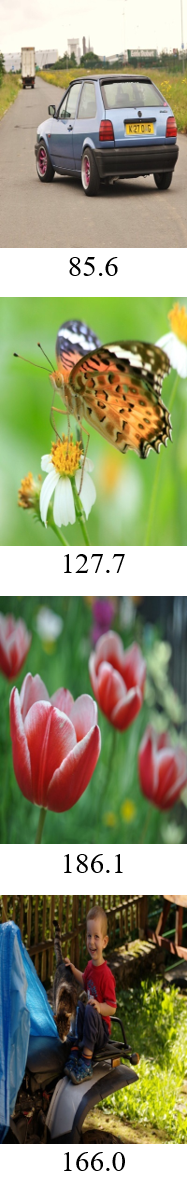}
          \vspace{-5ex}
          \subcaption{\footnotesize MULTIOP}
          \label{fig:qualitative_comparison:multiop}
          \end{center}
      \end{minipage}
      \begin{minipage}{0.105\hsize}
          \begin{center}
          \includegraphics[width=\hsize]{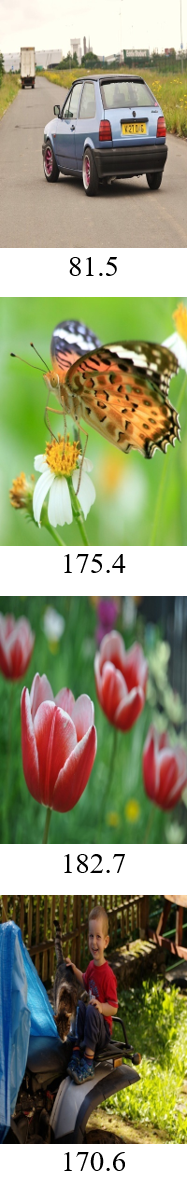}
          \vspace{-5ex}
          \subcaption{\footnotesize Ours}
          \label{fig:qualitative_comparison:ours}
          \end{center}
      \end{minipage}
      \begin{minipage}{0.105\hsize}
          \begin{center}
          \includegraphics[width=\hsize]{fig/result_scl_2.png}
          \vspace{-5ex}
          \subcaption{\footnotesize}
          \label{fig:qualitative_comparison:ablation1}
          \end{center}
      \end{minipage}
      \begin{minipage}{0.105\hsize}
          \begin{center}
          \includegraphics[width=\hsize]{fig/result_scl_2.png}
          \vspace{-5ex}
          \subcaption{\footnotesize}
          \label{fig:qualitative_comparison:ablation2}
          \end{center}
      \end{minipage}
  \end{tabular}
  \vspace{-2ex}
  \caption{Qualitative comparison with other methods where the retargeted size was 50\% of the original size  : (a) original image, (b) SCL, (c) GAIC~\cite{zeng2019reliable}, (d) SC~\cite{rubinstein2008improved}, (e) MULTIOP~\cite{rubinstein2009multi}, (f) Our method, (g) Our method without both the self-play-based reward and the frequency-aware weighted loss, (h) Our method without the frequency-aware weighted loss. The number below each image represents the BDW score. A lower score means a smaller distance from the original image. (\textcopyright David Locke, cattan2011, davemichuda, Honza Soukup)}
  \label{fig:qualitative_comparison}
\end{figure*}

\begin{figure}[t]
  \centering
  \begin{tabular}{c c c c}
      \begin{minipage}{0.38\hsize}
          \begin{center}
          \includegraphics[width=\hsize]{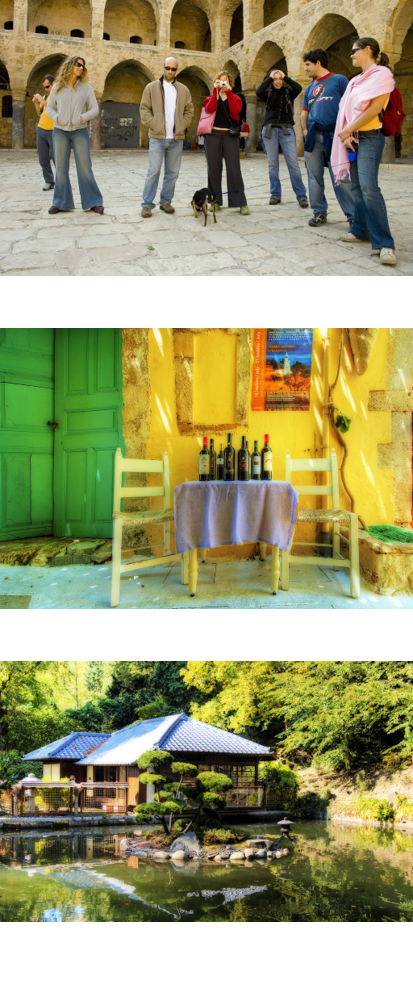}
          \vspace{-5ex}
          \subcaption{\footnotesize original image}
          \end{center}
      \end{minipage}
      \begin{minipage}{0.19\hsize}
          \begin{center}
          \includegraphics[width=\hsize]{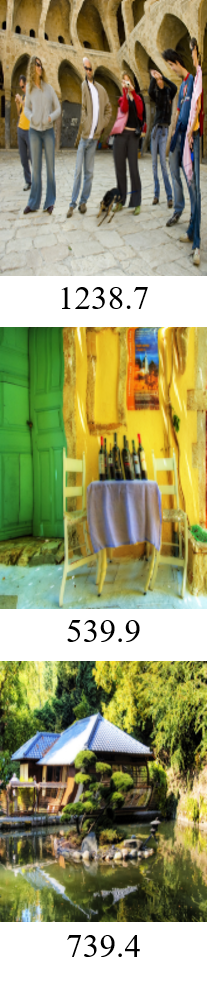}
          \vspace{-5ex}
          \subcaption{\footnotesize WSSDCNN}
          \end{center}
      \end{minipage}
      \begin{minipage}{0.19\hsize}
          \begin{center}
          \includegraphics[width=\hsize]{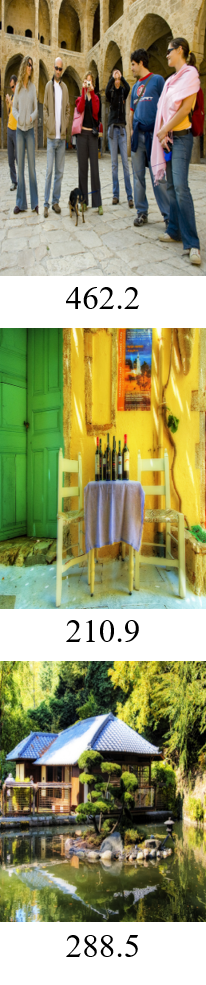}
          \vspace{-5ex}
          \subcaption{\footnotesize Cycle-IR}
          \end{center}
      \end{minipage}
      \begin{minipage}{0.19\hsize}
          \begin{center}
          \includegraphics[width=\hsize]{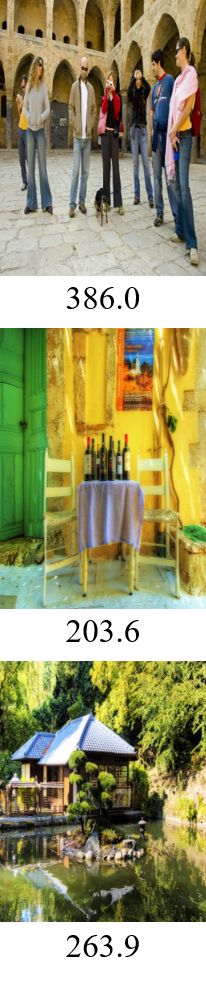}
          \vspace{-5ex}
          \subcaption{\footnotesize Ours}
          \end{center}
      \end{minipage}
  \end{tabular}
  \vspace{-2ex}
  \caption{Qualitative comparison with other deep-learning-based methods where the retargeted size was 50\% of the original size : (a) original image, (b) WSSDCNN~\cite{cho2017weakly}, (c) Cycle-IR~\cite{tan2019cycle}, (d) Our method. The number below each image represents the BDW score. (\textcopyright yanivba, Woflgang Staudt)}
  \label{fig:qualitative_comparison2}
\end{figure}

\begin{figure*}[t]
  \includegraphics[width=\textwidth]{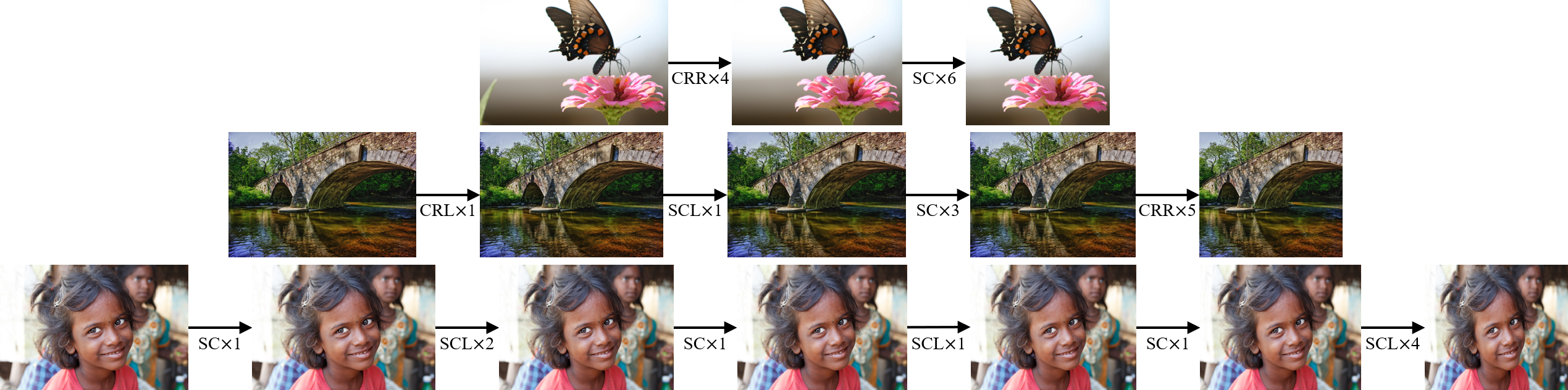}
  \caption{Examples of sequential actions selected by the agent. The symbol below the arrow indicates the applied retargeting operator and the number of times it is applied. (\textcopyright Lindsey Turner, Ted Van Pelt, Nithi Anand)}
  \label{fig:result_process}
\end{figure*}

\begin{figure*}[t]
  \centering
  \begin{tabular}{c c c c c c c c}
      \begin{minipage}{0.15\hsize}
          \begin{center}
          \includegraphics[width=\hsize]{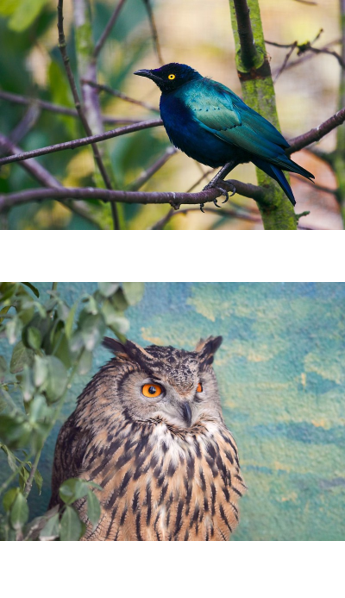}
          \vspace{-5ex}
          \subcaption{\footnotesize original image}
          \end{center}
      \end{minipage}
      \begin{minipage}{0.112\hsize}
          \begin{center}
          \includegraphics[width=\hsize]{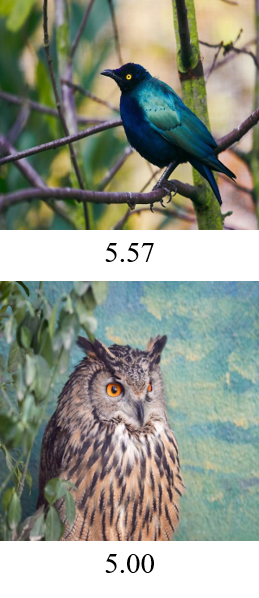}
          \vspace{-5ex}
          \subcaption{\footnotesize SCL}
          \end{center}
      \end{minipage}
      \begin{minipage}{0.112\hsize}
          \begin{center}
          \includegraphics[width=\hsize]{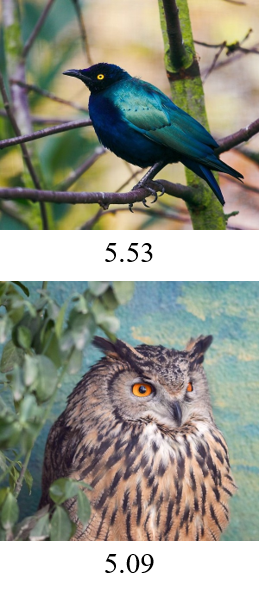}
          \vspace{-5ex}
          \subcaption{\footnotesize GAIC}
          \end{center}
      \end{minipage}
      \begin{minipage}{0.112\hsize}
          \begin{center}
          \includegraphics[width=\hsize]{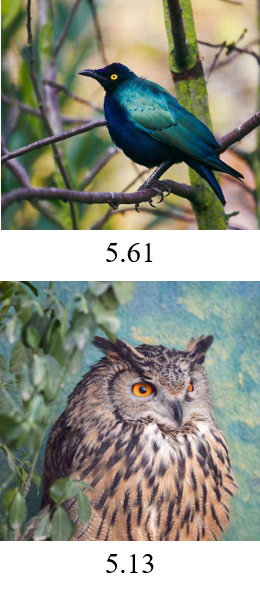}
          \vspace{-5ex}
          \subcaption{\footnotesize SC}
          \end{center}
      \end{minipage}
      \begin{minipage}{0.112\hsize}
          \begin{center}
          \includegraphics[width=\hsize]{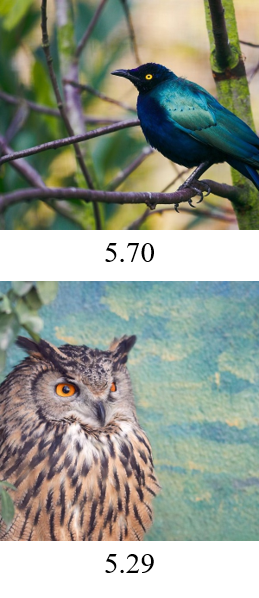}
          \vspace{-5ex}
          \subcaption{\footnotesize MULTIOP}
          \end{center}
      \end{minipage}
      \begin{minipage}{0.112\hsize}
          \begin{center}
          \includegraphics[width=\hsize]{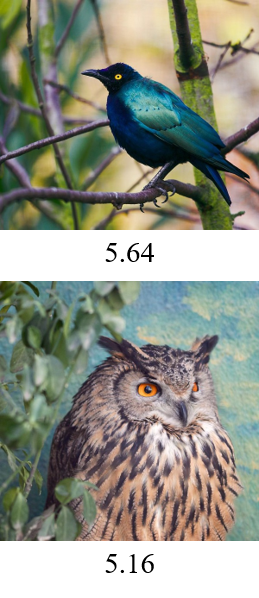}
          \vspace{-5ex}
          \subcaption{\footnotesize Ours}
          \end{center}
      \end{minipage}
      \begin{minipage}{0.112\hsize}
          \begin{center}
          \includegraphics[width=\hsize]{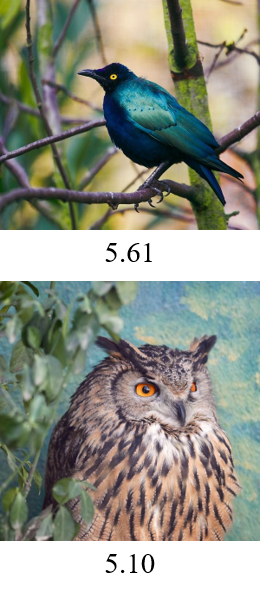}
          \vspace{-5ex}
          \subcaption{\footnotesize}
          \end{center}
      \end{minipage}
      \begin{minipage}{0.112\hsize}
          \begin{center}
          \includegraphics[width=\hsize]{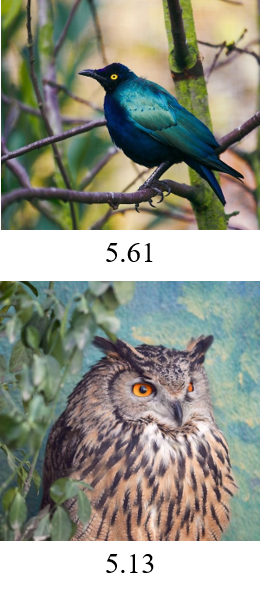}
          \vspace{-5ex}
          \subcaption{\footnotesize}
          \end{center}
      \end{minipage}
  \end{tabular}
  \vspace{-2ex}
  \caption{Qualitative comparison when the aesthetic score was used as a reward to generate 75\% retargeted images : (a) original image, (b) SCL, (c) GAIC~\cite{zeng2019reliable}, (d) SC~\cite{rubinstein2008improved}, (e) MULTIOP~\cite{rubinstein2009multi}, (f) Our method, (g) Our method without both the self-play-based reward and the frequency-aware weighted loss, (h) Our method without the frequency-aware weighted loss. The number below each image represents the aesthetic score; a higher score means a more aesthetic image. (\textcopyright Klaus Post, jennifer gergen)}
  \label{fig:qualitative_comparison3}
\end{figure*}

\section{Experiments}

\subsection{Experimental Settings}

\subsubsection*{Dataset}
\label{dataset}
To train and test our model, we used the MIRFLICKR-1M dataset~\cite{huiskes2010new}, 
which was composed of one million images downloaded from Flickr\footnote{https://www.flickr.com/} under the Creative Commons license.
% which was composed of one million images that were downloaded from Flickr\footnote{https://www.flickr.com/}.
From this dataset, we extracted 3,000 and 100 landscape-oriented images without loss of generality for training and testing, respectively.
As an additional test dataset, we used the RetargetMe dataset~\cite{rubinstein2010comparative},
which was the benchmark for image retargeting and contained 80 images.
We selected 68 landscape-oriented images for testing.

\subsubsection*{Implementation}
We used landscape-oriented images as inputs and retargeted them to shorten their width.
During training, $i_{max}$ was randomly sampled from $\{1,...,20\}$ at the start of the episode.
As the retargeted image was 2.5\% shorter than the original image in each step, the final retarget image size was 97.5\% to 50\% of the original size.
During the test process, by specifying $i_{max}$, retargeted images with the desired size were obtained.

During the observation, the original image was resized so that the width was 40 pixels; it was then padded with zero values so that the height was also 40 pixels.
Furthermore, the retargeted image was resized so that the height was the same as the resized original image; it was then padded with zero values so that the height was 40 pixels.
The vector ${\bf u}$, which represents the number of steps to the end of the episode, was set to 20 dimensions.
The elements of ${\bf u}$ from the first to the $(20-i_{max})$-th were initialized to zero; the other elements were initialized as one.
At the end of the $i$-th step, the $(20-i_{max}+i)$-th element was changed to zero.

Our model started with a 6-layer convolution block, which received the merged original and retargeted image and outputted a 25,600-dimensional vector.
The output was concatenated with the vector ${\bf u}$ and was the input for three fully-connected layers and an LSTM layer.
% The LSTM layer was used to memorize historical observations.
It outputted a 1,024-dimensional vector into the value network and policy network.
The value network consisted of one fully-connected layer and outputted the estimated value of the current state.
Likewise, the policy network consisted of one fully-connected layer but outputted a 4-dimensional vector, where each element corresponded to the probabilities for taking each action. 

We optimized our model utilizing the RMSProp~\cite{tieleman2012lecture} algorithm with a learning rate of $7\times 10^{-4}$; the other parameters were set to their default values.
We trained the networks for 10,000 episodes and used the model after the final episode for the test; it took 35 hours to complete the training.
The mini-batch size $|\mathcal{B}|$, the discount factor $\gamma$, and the weight of the entropy loss $\beta$ were set to $16$, $0.99$, and $0.01$, respectively.

\subsection{Qualitative Evaluation}
We show the retargeting results of our method and some previous methods in Figure~\ref{fig:qualitative_comparison} and Figure~\ref{fig:qualitative_comparison2}, where the retargeted size is 50\% of the original size.
The previous methods; scaling, GAIC~\cite{zeng2019reliable}, seam carving~\cite{rubinstein2008improved}, MULTIOP~\cite{rubinstein2009multi}, WSSDCNN~\cite{cho2017weakly}, and Cycle-IR~\cite{tan2019cycle} were compared.
The retargeting operators of scaling and seam carving were the same as the operators used in our method.
GAIC is a method used to find the optimal cropping window by taking into account the contents of the image.
MULTIOP uses multiple operators and searches for the optimal combination using dynamic programming; this increases the computational time.
WSSDCNN and Cycle-IR are deep-learning-based methods.
% WSSDCNN learns the shift map of each pixel in the input and output images, and Cycle-IR learns the forward and reverse mapping of the input and output images. 
As shown in Figure~\ref{fig:qualitative_comparison} and Figure~\ref{fig:qualitative_comparison2}, the retargeting results by scaling, seam carving, WSSDCNN, and Cycle-IR retained the information of the original image; nevertheless, the structure was distorted.
GAIC generated natural results; however, the information of the original image was lost.
Compared to these methods, MULTIOP preserved the natural structure of the original image and its important information.
Our method could also naturally retarget images while keeping important information; our results are very similar to the results of MULTIOP.
The BDW scores also show that the results from MULTIOP, and our method is almost the same and better than the other methods.
These results show that our method achieved a multi-operator image retargeting that has the same performance as MULTIOP.
Figure~\ref{fig:result_process} shows the results where the retargeted size was 75\% of the original size and how the actions were sequentially applied to the images.
It is shown that the appropriate combination of the retargeting operators for each image was selected by the agent and that the multi-operator image retargeting was achieved through reinforcement learning.

To analyze our method, we conducted ablation experiments.
We used a self-play-based reward to deal with the large variance in the BDW scores.
To verify the efficacy of the self-play-based reward, we conducted an experiment where the BDW score was given to the agent as a reward instead of our self-play-based reward.
In this experiment, the agent received the following reward every step,
\begin{equation}
    R_{ij} = {\rm BDW}(I_j^{orig}, I_{(i+1)j}) - {\rm BDW}(I_j^{orig},I_{ij}).
\end{equation}
The results are shown in Figure~\ref{fig:qualitative_comparison}(g).
Under this setting, the agent selected only scaling; the retargeted result is the same as those by scaling (Figure~\ref{fig:qualitative_comparison}(b)).
Owing to the value network not being able to approximate the reward, only a relatively strong action ({\it i.e.}, scaling) is selected for every image.

Moreover, to verify the efficacy of the frequency-aware weighted loss, we conducted an experiment where the training loss weight was fixed as $w_{ij} = 1$.
% Furthermore, to verify the efficacy of the frequency-aware weighted loss, we conducted an experiment where the loss weight was fixed as $w_{ij} = 1$ during the training process.
The results are shown in Figure~\ref{fig:qualitative_comparison}(h).
Under this setting, the agent only selected scaling again.
This is because the chances of victory increase by just picking a relatively strong action.
These results show that our two contributions are essential to achieve reinforcement-learning-based image retargeting.

\subsection{User Study}
We evaluated our method through a user study.
As described in~\ref{dataset}, we used 100 images from the MIRFLICKR-1M dataset~\cite{huiskes2010new} and 68 images from the RetargetMe dataset~\cite{rubinstein2010comparative}.
The test images of the MIRFLICKR-1M dataset were retargeted to both 75\% and 50\% of the original width. 
Regarding the RetargetMe dataset, 39 landscape-oriented images were retargeted to 75\% of the original width and 29 landscape-oriented images were retargeted to 50\% of the original width.
As Cycle-IR~\cite{tan2019cycle} only published retargeting results where the images from the RetargetMe dataset were retargeted to 50\% of the original width, we compared theirs with our results only under that condition.
In a user study, 50 crowd workers via the Amazon Mechanical Turk were asked to compare two images retargeted by our method and one of the previous methods; they were then instructed to select the better image.
All images were arranged randomly to avoid bias.
Table~\ref{tab:userstudy} shows the average vote rate.
Our method obtained a higher vote rate than scaling, GAIC~\cite{zeng2019reliable}, seam carving~\cite{rubinstein2008improved}, WSSDCNN~\cite{cho2017weakly}, and Cycle-IR~\cite{tan2019cycle}; this shows that our method is subjectively superior to these methods.
Furthermore, there was almost no difference in the average vote rate between the MULTIOP~\cite{rubinstein2009multi} and our method, showing that the performance of our method was subjectively similar to that of MULTIOP.

\newcolumntype{D}{>{\centering\arraybackslash}p{14.3mm}}
\begin{table}[t]
  \caption{Average vote rate$[\%]$.}
  \label{tab:userstudy}
  {\tabcolsep=0mm
  \begin{tabular}{lDDDD}
      \toprule
      \small
      & \multicolumn{2}{c}{75\%} & \multicolumn{2}{c}{50\%} \\
      Method & {\scriptsize RetargetMe} & {\scriptsize MIRFLICKR-1M} & {\scriptsize RetargetMe} & {\scriptsize MIRFLICKR-1M} \\
      \midrule
      SCL~/~Ours & 43.8/{\bf 56.2} & 42.1/{\bf 57.9} & 40.5/{\bf 59.5} & 46.2/{\bf 53.8} \\
      GAIC~\cite{zeng2019reliable}~/~Ours & 42.2/{\bf 57.8} & 45.7/{\bf 54.3} & 33.0/{\bf 67.0} & 37.8/{\bf 62.2} \\
      SC~\cite{rubinstein2008improved}~/~Ours & 40.9/{\bf 59.1} & 46.6/{\bf 53.4} & 42.3/{\bf 57.7} & 44.6/{\bf 55.4} \\
      MULTIOP~\cite{rubinstein2009multi}~/~Ours & 48.3/{\bf 51.7} & {\bf 50.2}/49.8 & 48.5/{\bf 51.5} & 48.7/{\bf 51.3}\\
      WSSDCNN~\cite{cho2017weakly}~/~Ours & 41.5/{\bf 58.5} & 45.2/{\bf 54.8} & 32.5/{\bf 67.5} & 40.8/{\bf 59.2}\\
      Cycle-IR~\cite{tan2019cycle}~/~Ours & - & - & 46.9/{\bf 53.1} & -\\
      \bottomrule
  \end{tabular}
  }
\end{table}

\subsection{Time Efficiency}
The benefit of the proposed method is that we can retarget images much faster than MULTIOP~\cite{rubinstein2009multi} while maintaining the image quality by using reinforcement learning.
To show the time efficiency of our method, we compared the computational time of our method to that of MULTIOP.
We used the RetargetMe dataset~\cite{rubinstein2010comparative} for the evaluation; all input images were resized to 640 $\times$ 480 px.
All processes were done on the same machine, which had an Intel$\text{\textregistered}$~Xeon$\text{\textregistered}$~Gold 6136~(3.00~GHz)~CPU.
The average computational time of our model and MULTIOP~\cite{rubinstein2009multi} are shown in Table~\ref{tab:computation_time}.
As shown in this table, our method achieved a multi-operator image retargeting that was faster by three orders of magnitude than MULTIOP.
Compared to MULTIOP, which generated multiple images and evaluated them with BDW, the proposed method predicted the appropriate operators step by step, which resulted in a faster image retargeting.
The time and space complexities of the MULTIOP~\cite{rubinstein2009multi} were $O(i^n)$.
As this is a polynomial in the number of steps $i$, but an exponential in the number of operators $n$, the calculation time of MULTIOP increased greatly as the number of steps gets bigger.
In comparison, the computational time of our method did not change drastically, even when the retarget ratio increased.
This trend became more pronounced as the number of operators increased.

\begin{table}
  \caption{Average computational time of 640 $\times$ 480 images.}
  \label{tab:computation_time}
  \begin{tabular}{lcc}
    \toprule
    Method & 75\% & 50\% \\
    \midrule
    MULTIOP~\cite{rubinstein2009multi} & 3400 s & 32000 s\\
    Ours & 5.0 s & 9.9 s \\
  \bottomrule
\end{tabular}
\end{table}

\subsection{Reward Option}
In the above sections, the BDW scores are used as the reward, and our model searches for a retargeted image that minimizes the BDW score.
In addition to the BDW score, other reward functions can be integrated into our method very easily.
In this section, we show the experiment where the reward is replaced with another evaluation function, the aesthetic score proposed by Wang et al.~\cite{wang2019aspect}.
In this setting, we search for a retargeted image that maximized the aesthetic score.
When the aesthetic function is denoted as ${\rm AES}$, the reward Eq. (\ref{eq:sp-reward}) is replaced as follows,
\begin{equation}
  R_{i_{max}j}=\left\{
  \begin{array}{ll}
    \!\!+1~({\rm if~} {\rm AES}(I_{i_{max}j})\!>\!{\rm AES}(I'_{i_{max}j}))\!\!\!\!\\
    \!\!-1~({\rm otherwise})
  \end{array}.
  \right.
\end{equation}

Figure~\ref{fig:qualitative_comparison3} shows the results of our model based on the aesthetic score and other methods.
For comparison, we conduct experiments where MULTIOP~\cite{rubinstein2009multi} optimized the aesthetic score instead of the BDW score (Figure~\ref{fig:qualitative_comparison3}(e)).
% Our method (Figure~\ref{fig:qualitative_comparison3}(f)) generated retargeted images that were not much different from those generated by MULTIOP~\cite{rubinstein2009multi}; furthermore, it obtained higher aesthetic scores than the other methods previously described.
Although our retargeting results (Fig~\ref{fig:qualitative_comparison3}(f)) did not exactly match results by MULTIOP~\cite{rubinstein2009multi}, our method obtained higher aesthetic scores than the other methods previously described.
These results show that the efficacy of our method does not depend on the type of evaluation function.

\section{Conclusions}
In this study, we addressed image retargeting, a task where we adjust input images into arbitrary sizes. 
Despite the previous multi-operator method having a high performance, the method required a huge computational time for generating multiple retargeted images to find the best combination of retargeting operators.
Therefore, we proposed a reinforcement-learning-based method to achieve fast multi-operator image retargeting by predicting the optimal retargeting operator step by step.
To deal with issues of a large variance in the evaluation value, and a local optimum where only the relatively strong action is selected, we proposed a self-play-based reward and a frequency-aware weighted loss.
These two contributions enabled us to achieve a fast and effective multi-operator image retargeting via reinforcement learning.
Experimental results showed that our method achieved multi-operator image retargeting that was faster by three orders of magnitude and had the same performance as the state-of-the-art method.

\begin{acks}
A part of this research was supported by JSPS KAKENHI Grant Number 18H03339, 19K20289.
\end{acks}

\bibliographystyle{ACM-Reference-Format}
\bibliography{reference}

\end{document}